\begingroup\expandafter\expandafter\expandafter\endgroup
\expandafter\ifx\csname XeTeXversion\endcsname\relax
  \pdfoutput=1
\fi
\documentclass[10pt,letterpaper,twocolumn]{article}
\usepackage[top=0.75in,bottom=0.75in,left=0.75in,right=0.75in,columnsep=0.25in]{geometry}
\usepackage[T1]{fontenc}
\usepackage{newtxtext,newtxmath}
\usepackage{amsmath}
\usepackage{graphicx}
\usepackage[hyphens]{url}
\usepackage{natbib}
\usepackage{caption}
\usepackage{algorithm}
\usepackage{algorithmic}
\usepackage{newfloat}
\usepackage{listings}
\usepackage{booktabs}

\DeclareCaptionStyle{ruled}{labelfont=normalfont,labelsep=colon,strut=off}
\floatstyle{ruled}
\newfloat{listing}{tb}{lst}{}
\floatname{listing}{Listing}
\title{Channel Match: Calibrating Shape Latents for\\ Few-Step 3D Generation}
\author{Di Yang$^{1}$\qquad Zhihao Li$^{2}$\qquad
Yanhai Xiong$^{1}$\qquad Yufei Wang$^{2}$\\[6pt]
$^{1}$College of William \& Mary\qquad $^{2}$SparcAI Inc.\\[6pt]
{\small\ttfamily dyang06@wm.edu\qquad zhihao.li@sparclab.ai}\\
{\small\ttfamily yxiong05@wm.edu\qquad yufei.wang@sparclab.ai}}
\date{}

\begin{document}
\maketitle

\begin{abstract}
Recent native 3D generators achieve high-quality asset creation by modeling structured 3D latents with flow-based transformers, but their high-resolution generation remains computationally expensive due to costly iterative sampling. In this work, we introduce a few-step distillation framework for native 3D generation, enabled by continuous self-consistency training with a custom JVP FlashAttention kernel and an analytic velocity regression objective for sparse structured 3D latents. To stabilize joint optimization, we further introduce an adaptive loss weighting strategy that automatically balances them without manual tuning. Beyond acceleration, we reveal a previously overlooked failure mode of few-step native 3D generation: sampling-induced latent distribution drift preserves geometric correctness but disrupts downstream PBR texture synthesis, causing texture artifacts on geometrically valid assets. To address this issue, we propose Channel Match, a training-free adaptive per-channel latent calibration method that restores the latent distribution without modifying either generator. We validate our approach on TRELLIS.2 at the $1024^3$ resolution setting. Our two-step model achieves $4.5\times$ faster geometry sampling while maintaining geometry quality comparable to the many-step teacher, and Channel Match further improves rendering quality by recovering the latent interface between geometry and appearance generation.
\end{abstract}

\begin{figure*}[htbp]
  \centering
  \includegraphics[width=1\textwidth]{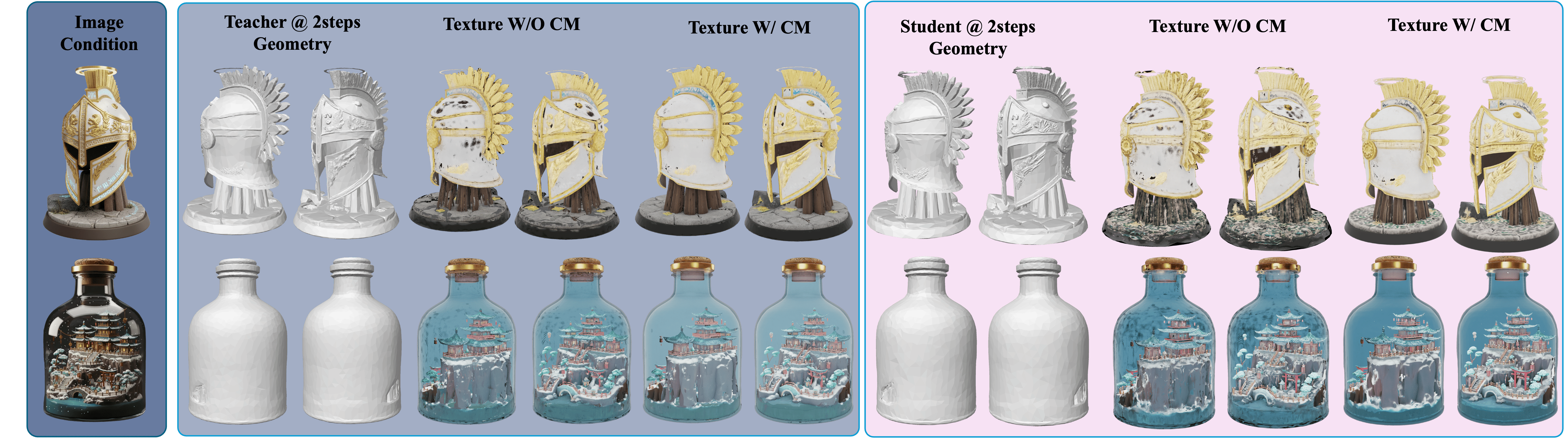}
\caption{The few-step distribution shift and its correction by \textbf{CM}. From left: input image; the untextured mesh and PBR render of our two-step student without Channel Match, where the sampling-induced shifted shape latent drives the frozen PBR texture model to produce faulty surface materials (dark/blotchy patches); and the mesh and render with \textbf{CM}. For both few-step teacher and student sampling, calibrating the latent removes the texture artifacts, while the geometry remains unchanged. (We consider samples with simple geometry here, where both the two-step teacher and our two-step student already recover structures close to the many-step reference. This controlled setting isolates the effect of latent distribution shift rather than geometry reconstruction errors, showing that the shift is a general issue of few-step sampling and can be effectively corrected by \textbf{CM}.)}
  \label{fig:teaser}
\end{figure*}
\section{Introduction}

Recent advances in native 3D generators have enabled high-quality asset creation by directly modeling structured 3D representations. These models encode assets into sparse and structured latent spaces and generate them with flow-based transformers~\citep{xiang2024trellis,liu2023rectified,lipman2023flow}. Despite their impressive details, high-resolution native 3D generators remain computationally expensive, as each flow step requires processing a large number of 3D tokens in structured latent spaces, and generation typically requires many iterative sampling steps. This computational cost limits their deployment in interactive applications.

Few-step distillation provides a promising direction to reduce this cost. Consistency models and their continuous-time formulations compress multi-step samplers into one or two steps~\citep{song2023consistency,lu2025scm}, while distribution-matching and adversarial distillation methods achieve near-teacher quality with even fewer evaluations~\citep{yin2024dmd,lin2024sdxllightning,sauer2024add}. However, directly applying these techniques to native 3D generation remains challenging due to fundamental differences in latent representations and generation pipelines.

First, native 3D generators rely on sparse and variable-length structured latents, whereas existing continuous-time distillation methods are primarily designed for dense fixed-dimensional representations~\citep{song2023consistency,lu2025scm}. Such latent structures are essential for scaling 3D generation to high resolutions, but introduce challenges for computing consistency objectives, particularly the velocity directional derivatives required by continuous-time formulations. Second, native 3D generation pipelines contain intermediate latent interfaces shared across multiple generation stages. Unlike image generation where the latent is directly decoded into pixels, a geometry latent in native 3D generation can serve as the input to downstream appearance synthesis, such as PBR texture generation. Therefore, reducing sampling steps may preserve geometric structure while shifting latent statistics away from the distribution expected by downstream generators, causing quality degradation that is invisible to geometry-only evaluation.

In this work, we investigate few-step distillation for high-resolution native 3D generation. We develop a distillation framework tailored for sparse structured 3D latents, enabling continuous self-consistency training through a custom JVP FlashAttention kernel. To avoid the ambiguity of learning a consistent but incorrect transport field, we incorporate an analytic velocity regression objective that anchors the student toward the true marginal transport. Furthermore, we introduce an adaptive loss weighting strategy that automatically balances self-consistency and velocity regression objectives across timesteps without manual tuning of loss weights.


Beyond acceleration, we reveal a previously overlooked failure mode of few-step 3D generation: sampling-induced latent distribution drift. The accelerated geometry sampler can produce structurally valid shape latents while deviating from the training distribution in channel statistics. When consumed by the downstream PBR texture generation stage, these shifted latents lead to incorrect material synthesis, producing surface artifacts on otherwise geometrically valid assets. To address this issue, we propose Adaptive Channel Match, a training-free adaptive per-channel latent calibration method that restores sampling-shifted latents toward the training distribution without modifying either generator.

We validate our approach on TRELLIS.2~\citep{xiang2025trellis2} at the $1024^3$ resolution setting. Our two-step model achieves $4.5\times$ faster geometry sampling while maintaining geometry quality comparable to the many-step teacher. Adaptive channel match further improves rendering quality by recovering the latent interface between geometry generation and appearance synthesis.

Our contributions are summarized as follows.
\begin{itemize}
  \item We reveal a previously overlooked failure mode in few-step native 3D generation: aggressive sampling reduction can introduce distribution shifts in the geometry latent that preserve geometric completeness but violate the latent interface with downstream appearance synthesis, causing texture artifacts that resemble structural defects.

  \item We propose Channel Match, a training-free adaptive per-channel latent calibration method that restores the shifted geometry latent toward the training distribution. Without retraining either the geometry or texture generator, it eliminates texture artifacts, improves paired rendering quality toward the many-step teacher, and mitigates similar degradation in naively few-step sampling.

  \item We develop a scalable few-step distillation framework for high-resolution native 3D flow models. By enabling continuous self-consistency training on sparse structured 3D latents with a custom JVP FlashAttention kernel and an analytic velocity regression objective, our two-step student achieves $4.5\times$ faster geometry sampling with quality comparable to the many-step teacher and surpassing SOTA baselines.
\end{itemize}

\section{Related Work}

\paragraph{Native 3D generation.}
Structured-latent models generate features on sparse active voxels via rectified-flow transformers, decoding to meshes, Gaussians, or radiance fields~\citep{xiang2024trellis,ren2024xcube,kerbl20233dgs}. TRELLIS.2 extends this line to $1024^3$-grid latents with PBR materials and a separate texture stage~\citep{xiang2025trellis2}. High-resolution sparse representations similarly enable detailed geometry at large grid sizes~\citep{li2026sparc3d,he2025sparseflex,clayzhang}. Other native generators diffuse compact vector-set or primitive latents~\citep{lai2025flashvdm,li2025triposg,chen20243dtopiaxl,zhao2025hunyuan3d2,chen2025dora}, and earlier feed-forward or latent-field methods trade fidelity for speed~\citep{tang2024lgm,lan2024ln3diff}. Supported by large datasets~\citep{deitke2023objaverse,deitke2023objaversexl}, we fix the shape and texture stages of TRELLIS.2 and distill only the shape flow, so any measured change stems from the shape latent alone.

\paragraph{Few-step distillation.}
Progressive distillation halves the step count
repeatedly~\citep{salimans2022progressive}; consistency models regress points on a trajectory to a shared endpoint~\citep{song2023consistency}, and the continuous-time
form removes the discretization schedule once its training instability is controlled through a simplified parameterization and an adaptive weighting~\citep{lu2025scm,karras2022edm}. A second family replaces trajectory matching with a distribution-level objective: distribution-matching, adversarial, and unified trajectory-distribution distillations~\citep{yin2024dmd,yin2024improved,sauer2024add,lin2024sdxllightning,luo2025tdm}, which unifies the two, reach one-to-few-step image generation. These methods rely on 2D image discriminators, whose signal is hard to obtain for a native 3D latent because quality evaluation requires non-differentiable rendering. We therefore use continuous-time self-consistency in velocity space, paired with a ground-truth velocity loss to anchor the fixed point.

\paragraph{Forward-mode attention and 3D consistency.}
The continuous objective needs a forward-mode tangent through attention. FlashAttention~\citep{dao2022flashattention} and its sparse extension SCFA~\citep{pagliardini2023faster} keep attention memory-efficient but expose only the primal output. The closest 3D consistency model, ConTiCoM-3D~\citep{eilermann2025conticom}, avoids Jacobian-vector products entirely because they are expensive for unordered point clouds, and works teacher-free in raw point space. 
Ours is the opposite: a teacher-based, latent-space, sparse-voxel pipeline where a custom fused kernel makes the Jacobian-vector product tractable at high resolution.

\paragraph{Accelerating 3D generation.}
Direct transfer of 2D distillation to native 3D is uncommon. FlashVDM accelerates both the sampler and the decoder of a vector-set model but generates shape without appearance~\citep{lai2025flashvdm}. A parallel line instead cuts attention cost with sparse attention~\citep{chen2025ultra3d,wu2025directds}. The marginal-data-transport distillation of \citet{zhou2025mdtdist} converts the transport objective into velocity matching and velocity distillation and reaches one-to-two-step generation on an earlier structured-latent backbone; it is the closest baseline, and we reimplement it on TRELLIS.2 for a controlled comparison. Neither line reports the latent distribution-shift failure that Channel Match targets, which appears once the shape stage is pushed to very few steps.

\paragraph{Statistics alignment.}
Matching per-channel mean and variance to a reference is a long-standing idea, from batch normalization~\citep{ioffe2015batchnorm} to adaptive instance normalization for style transfer~\citep{huang2017adain}. Channel Match uses the same primitive but contributes a diagnosis, not a layer: a few-step shape latent drifts off the training marginals, and realigning its per-channel statistics restores a frozen downstream stage.

\section{Preliminaries}

\subsection{Structured-Latent Generation in TRELLIS.2}
An asset is encoded as a sparse structured latent
$x = \{(x_i, p_i)\}_{i=1}^{L}$, where $p_i$ indexes an active voxel of a
high-resolution grid and $x_i \in \mathbb{R}^{C}$ is its feature. Generation
factorizes into three conditional flow stages driven by the input image $c$: a
sparse-structure stage predicts the active voxels; a shape stage generates the
geometry latent on those voxels; and a texture stage generates the PBR material
latent. A decoder $\mathcal{D}$ maps the shape and texture latents to a textured
mesh. The shape stage is coarse-to-fine: a low-resolution ($512$) flow produces an
initial latent whose active coordinates are upsampled to the $1024$ grid, and a
high-resolution ($1024$) flow generates the final geometry latent on those
coordinates. We distill this $1024$ shape flow. The sparse-structure stage, the 
$512$ flow that seeds the coordinates, and the texture stage are frozen and shared across all configurations leaving only the $1024$ shape flow to differ between teacher and student.

\subsection{Rectified Flow}
Let $x_0$ be a data latent and $\epsilon \sim \mathcal{N}(0, I)$ noise. Rectified
flow interpolates along a straight path
\begin{equation}
  x_t = (1-t)\,x_0 + \big(\sigma_{\min} + (1-\sigma_{\min})\,t\big)\,\epsilon,
  \qquad t \in [0,1],
  \label{eq:interp}
\end{equation}
with a small $\sigma_{\min}$, so $t=0$ is data and $t=1$ is noise. The instantaneous
velocity of this path is
\begin{equation}
  u(x_t,t) = \frac{\mathrm{d}x_t}{\mathrm{d}t} = (1-\sigma_{\min})\,\epsilon - x_0 .
  \label{eq:gtvel}
\end{equation}
A flow model $v_\phi$ is trained to regress the conditional expectation
$\bar v(x_t,t)=\mathbb{E}[\,u \mid x_t\,]$, which defines the probability-flow ODE
$\dot x_t = \bar v(x_t,t)$. Samples are drawn by integrating this ODE from $t=1$ to
$t=0$; the teacher uses a $K_\phi$-step Euler schedule, and our goal is a student
$v_\theta$ whose $K$-step Euler integral, with $K\in\{1,2,4\}$, matches the
teacher's output.

\subsection{Continuous-Time Self-Consistency}
The consistency function $f(x_t,t)$ maps any point on a probability-flow trajectory
to that trajectory's data endpoint~\citep{song2023consistency}. Consistency holds
when $f$ is constant along the trajectory, i.e.\ its total time derivative vanishes,
$\tfrac{\mathrm{d}}{\mathrm{d}t}f(x_t,t)=0$. Expressed through the velocity, this is
a differential constraint that a single-step map must satisfy at every $t$. The
continuous-time formulation of \citet{lu2025scm} makes this constraint trainable by
a simplified parameterization and an adaptive per-timestep weighting; we adopt it in
velocity space, which matches the rectified-flow backbone, and combine it with
Equation~\ref{eq:gtvel}.

\section{Method}

We distill the many-step teacher $v_\phi$ into a few-step student $v_\theta$,
initialized from the teacher, with two velocity-space losses that constrain
different properties of the field and a per-timestep weighting that balances them.

\subsection{Ground-Truth Velocity Regression}
The first loss regresses the student velocity onto the analytic target of
Equation~\ref{eq:gtvel},
\begin{equation}
  \mathcal{L}_{\mathrm{gt}} = \mathbb{E}_{x_0,\epsilon,t}\big[\,
    \big\| v_\theta(x_t,t) - \big((1-\sigma_{\min})\epsilon - x_0\big) \big\|_2^2 \,\big].
  \label{eq:lgt}
\end{equation}
Its minimizer is the true marginal field, so it fixes the values the student should
take, and expressing it in velocity space avoids the small-$t$ weighting pathology of
a sample-space reconstruction. This loss alone does not yield a few-step sampler; it anchors the next loss away from a consistent but incorrect transport.

\subsection{Continuous Self-Consistency}
The second loss makes the field invariant to the number of integration steps. Along
a teacher trajectory, the total time derivative of the student velocity is
\begin{equation}
  g(x_t,t) = \frac{\mathrm{d}}{\mathrm{d}t}\,v_\theta(x_t,t)
  = \partial_t v_\theta \;+\; \big(\nabla_x v_\theta\big)\,\bar v_\phi(x_t,t),
  \label{eq:tangent}
\end{equation}
where the transport direction is the teacher velocity $\bar v_\phi$. We evaluate
Equation~\ref{eq:tangent} with one forward-mode (JVP) pass: the dual tangent of the
latent is seeded with $\bar v_\phi$, giving the spatial term $(\nabla_x
v_\theta)\bar v_\phi$, and the dual tangent of the time input is seeded so that the
pass also returns $\partial_t v_\theta$. Concretely, the backbone consumes a scaled
time $1000\,t$, and seeding its tangent with the constant $1000$ yields
$\partial_t v_\theta$ in the same pass. Omitting the time term, or seeding the
spatial term with anything other than the teacher velocity, biases the tangent.

We normalize the per-token tangent with a smooth, magnitude-preserving robust cap.
With $r = \lVert g\rVert / (\kappa\cdot \mathrm{med}\lVert g\rVert)$ per token and a
clip factor $\kappa$,
\begin{equation}
  \hat g = \frac{\tanh(r)}{r}\; g .
  \label{eq:tannorm}
\end{equation}
As $r\!\to\!0$ the scale tends to $1$(small tangents preserved, correction vanishing once $g\!\to\!0$); as $r\!\to\!\infty$ the token norm saturates near $\kappa\cdot\mathrm{med}\lVert g\rVert$. Unlike unit-norm normalization, this keeps small consistency signals intact. The self-consistency target regresses the student toward a stop-gradient,
exponential-moving-average (EMA) copy $\tilde v_\theta$ of itself, moved along the
normalized tangent,
\begin{equation}
  v^\star = \operatorname{sg}\!\big(\tilde v_\theta(x_t,t) - \eta\,\omega\,t\,\hat g\big),
  \label{eq:scmtarget}
\end{equation}
\begin{equation}
  \mathcal{L}_{\mathrm{scm}} = \mathbb{E}\big[\,\| v_\theta(x_t,t) - v^\star \|_2^2 \,\big],
  \label{eq:lscm}
\end{equation}
where $\eta$ is a step scale and $\omega$ a warm-up factor ramped from $0$ to $1$ over the
first training steps. Anchoring the value to the EMA student rather than the teacher
avoids a biased fixed point that arises when a slightly mismatched teacher value is
used as both the target and the transport direction.

\subsection{Adaptive Loss Balancing}
The two losses differ in scale and vary with $t$. Rather than tune a per-$t$ schedule,
we apply an uncertainty weighting: for a per-sample loss $\ell$ and a learnable scalar
$w(t)$, the weighted term is $\ell\,e^{-w(t)}+w(t)$, whose minimizer
$w^\star(t)=\log\ell$ self-normalizes the term across
time~\citep{karras2022edm,lu2025scm}. We realize $w(\cdot)$ as a small Fourier-feature
multilayer perceptron trained jointly, and apply it to both terms, giving
\begin{equation}
  \mathcal{L} = \mathbb{E}\big[\,e^{-w_{\mathrm{scm}}(t)}\,\ell_{\mathrm{scm}} + w_{\mathrm{scm}}(t)
  \;+\; e^{-w_{\mathrm{gt}}(t)}\,\ell_{\mathrm{gt}} + w_{\mathrm{gt}}(t)\,\big].
  \label{eq:total}
\end{equation}
The reported models use the two velocity-space losses of
Equation~\ref{eq:total}.

\subsection{JVP FlashAttention for Sparse Latents}
Equation~\ref{eq:tangent} is a Jacobian-vector product, so evaluating it requires
forward-mode differentiation through every layer of the transformer, including
attention. Stock FlashAttention~\citep{dao2022flashattention} returns the output but
not its forward-mode tangent, and would forfeit its memory savings on the high-resolution sparse latent. We implement a custom JVP FlashAttention kernel that, given the primal and tangent of the packed query-key-value tensor of the variable-length sparse latent, returns the primal output and its tangent in a single fused pass, and we patch the variable-length attention entry points used by the backbone to call it during the tangent computation. The kernel keeps the tangent evaluation near the memory cost of standard attention (Section~\ref{sec:jvpcost}), making the continuous objective usable at high resolution. We give the full kernel implementation in the supplementary material. Algorithm~\ref{alg:main} summarizes one training step.
\begin{algorithm}[htbp]
\caption{Few-step shape distillation and generation}
\label{alg:main}
\textbf{Training (one step).} \textbf{Input}: teacher $v_\phi$, student $v_\theta$,
EMA student $\tilde v_\theta$, weighting net $w$; clip $\kappa$, step $\eta$, warm-up $\omega$
\begin{algorithmic}[1]
\STATE Draw a token-balanced micro-batch; load latents $x_0$, conditions $c$
\STATE $\epsilon\!\sim\!\mathcal{N}(0,I)$, $t\!\sim\!\mathcal{U}(0,1)$; form $x_t$ by Eq.~\ref{eq:interp}
\STATE $\bar v \leftarrow v_\phi(x_t,t,c)$ \hfill// teacher direction, no grad
\STATE $g \leftarrow \partial_t v_\theta + (\nabla_x v_\theta)\,\bar v$ via JVP (dual seeds $\bar v$, $1000$)
\STATE $\hat g \leftarrow (\tanh r / r)\,g$, \ $r=\lVert g\rVert/(\kappa\,\mathrm{med}\lVert g\rVert)$
\STATE $\tilde v \leftarrow \tilde v_\theta(x_t,t,c)$; \ \ $v \leftarrow v_\theta(x_t,t,c)$
\STATE $v^\star \leftarrow \operatorname{sg}(\tilde v - \eta\,\omega\,t\,\hat g)$; \ \ $v_{\mathrm{gt}} \leftarrow (1-\sigma_{\min})\epsilon - x_0$
\STATE $\ell_{\mathrm{scm}} \leftarrow \lVert v - v^\star\rVert_2^2$; \ \ $\ell_{\mathrm{gt}} \leftarrow \lVert v - v_{\mathrm{gt}}\rVert_2^2$
\STATE $\mathcal{L} \leftarrow$ uncertainty-weighted sum (Eq.~\ref{eq:total}); backprop; update $v_\theta$, $w$; EMA update $\tilde v_\theta$
\end{algorithmic}
\vspace{0.4em}
\textbf{Generation with Channel Match.} \textbf{Input}: image $c$; frozen prefix,
texture flow, decoder $\mathcal{D}$; distilled $v_\theta$; stats $(\hat\mu,\hat\sigma)$; threshold $\tau_S$; steps $K$
\begin{algorithmic}[1]
\STATE $\mathcal{S} \leftarrow$ sparse structure$(c)$; \ upsample to $1024$ coords \hfill// shared prefix
\STATE $x \leftarrow$ $K$-step Euler integral of $v_\theta$ from noise on those coords
\STATE $S \leftarrow \frac{1}{C}\sum_c \max(0,\ \hat\sigma_c/\sigma_c^{\text{s}}(x)-1)$ \hfill// Eq.~\ref{eq:collapse}
\IF{$S > \tau_S$}
  \STATE $x \leftarrow$ ChannelMatch$(x;\hat\mu,\hat\sigma)$ \hfill// Eq.~\ref{eq:cm}
\ENDIF
\STATE $y \leftarrow$ texture flow$(x, c)$; \ \ $M \leftarrow \mathcal{D}(x, y)$; \ \textbf{return} $M$
\end{algorithmic}
\end{algorithm}

\subsection{Channel Match}
\label{sec:cm}
After distillation the decoded mesh is geometrically complete, but the
generated shape latent moves off the training distribution in its per-channel
value statistics, and the frozen PBR texture stage, trained only on
in-distribution latents, produces incorrect materials in the affected regions
(Figure~\ref{fig:teaser}). The geometry is intact, only the appearance degrades. Channel Match calibrates the shape latent to remove this shift before it reaches the texture stage.

For a latent $x=\{(x_i,p_i)\}_{i=1}^{L}$ with $x_i\in\mathbb{R}^{C}$, we treat
each channel $c$ independently. The stored training statistics
$(\hat\mu_c,\hat\sigma_c)$ are the mean and standard deviation of channel $c$
over all active tokens of the training set, computed once offline; the sampled
statistics $(\mu_c^{\text{s}},\sigma_c^{\text{s}})$ are the same two quantities
computed over the active tokens of the current generated latent. With a
strength $s$, the calibrated feature is
\begin{equation}
  x'_c = \hat\mu_c + (x_c - \mu_c^{\text{s}})\Big(1 + s\big(\hat\sigma_c/\sigma_c^{\text{s}} - 1\big)\Big).
  \label{eq:cm}
\end{equation}
Only dataset-level statistics enter, so the calibration is training-free and
modifies neither model. It is a per-channel affine map and does not model
inter-channel correlation, which we revisit in the limitations. We evaluate two
forms, full ($s{=}1$, matching mean and std) and std-only (keeping the asset's
mean); a mean-only variant is not considered, since the observed drift is a collapse of the per-channel standard deviation (Table~\ref{tab:drift}).

The shift depends on the step count, so we gate the calibration by a collapse
score computed from the sampled latent alone,
\begin{equation}
  S = \frac{1}{C}\sum_{c}\max\!\Big(0,\ \frac{\hat\sigma_c}{\sigma_c^{\text{s}}} - 1\Big),
  \label{eq:collapse}
\end{equation}
and apply it only when $S$ exceeds a threshold $\tau_S$, so no ground truth is
needed at inference. Algorithm~\ref{alg:main} places Channel Match as a single
in-place step between the distilled shape flow and the frozen texture flow; the
shared prefix and the texture stage are unchanged.

\section{Experiments}
\label{sec:Exp}
\subsection{Implementation Details}
The student is initialized from the pretrained TRELLIS.2-4B shape model, and we
distill only the $1024$-resolution shape flow; the sparse-structure stage, the $512$
shape flow, and the texture stage are frozen. Training runs on eight A800-80GB GPUs
with distributed data parallelism. All experiments use a fixed random seed of 42 for reproducibility. Because sparse latents have a variable number of
active voxels, we don't use a fixed batch size; instead we pack assets into each step
until a token budget of $40{,}000$ active tokens is reached and accumulate gradients,
balancing per-step compute across differently sized assets. We use Adam with learning rate $2\times10^{-6}$, $(\beta_1,\beta_2)=(0.0,0.999)$, and gradient
clipping at $1.0$; the weighting network uses learning rate $10^{-2}$. The EMA decay is
$0.9999$, $\sigma_{\min}=10^{-5}$, the tangent clip is $\kappa=4$ times the running
median, the tangent warm-up is $1000$ steps, and the step scale is $\eta=1$; the two
losses are equally weighted and training runs up to $80{,}000$ steps in bfloat16. The
JVP FlashAttention kernel is a fused Triton kernel that patches the backbone's
variable-length attention during the tangent pass. The channel statistics
$(\hat\mu,\hat\sigma)$ for Channel Match are estimated once on the training latents.

We compare our student at one, two, and four steps against the pretrained teacher run
at twelve steps (a many-step upper bound) and at two steps (a naive few-step baseline),
the marginal-data-transport distillation of \citet{zhou2025mdtdist} reimplemented on
this backbone (MDT), and a self-consistency-only ablation of our method that drops the
ground-truth loss $\mathcal{L}_{\mathrm{gt}}$ (SCM). All configurations share the same
frozen prefix and texture stage.

\subsection{Datasets}
We distill on a $100$K-asset subset of Objaverse and
Objaverse-XL~\citep{deitke2023objaverse,deitke2023objaversexl}, each asset providing a
ground-truth shape latent and a rendered conditioning image. For evaluation we use all
$3{,}216$ assets of the Toys4K object set~\citep{stojanov2021toys4k}, each conditioned
on a single rendered view.

\subsection{Metrics}
Geometry and appearance are scored under a fixed textured-mesh protocol: four views at
$30^\circ$ elevation and azimuths $\{0,90,180,270\}$. We report set-level Fréchet
distance in Inception~\citep{szegedy2016inception,heusel2017fid} and
DINOv2~\citep{oquab2024dinov2,stein2023exposing} feature spaces over the renders and
ULIP-2 image--shape cosine similarity~\citep{xue2024ulip2} averaged over views.
We render appearance from the PBR-textured mesh that TRELLIS.2 produces. The starred baselines in Table~\ref{tab:main} use varying representations (Gaussians, neural fields, primitive sets) and evaluation protocols. They serve only as loose references. We draw controlled conclusions strictly from the lower block, where all configurations share the identical frozen prefix, texture stage, and render protocol. \textbf{CM} acts locally, so a set-level Fréchet distance is the wrong instrument for it; we instead take the many-step teacher's render as the reference and report paired LPIPS~\citep{zhang2018lpips}, SSIM~\citep{wang2004ssim}, and PSNR between each student render and the teacher render under the same protocol, with the fraction of assets each calibration moves toward the teacher. Timing is measured on a single A800-80GB GPU.

\subsection{Distillation: Geometry and Efficiency}
\begin{figure*}[t]
  \centering
  \includegraphics[width=0.86\textwidth]{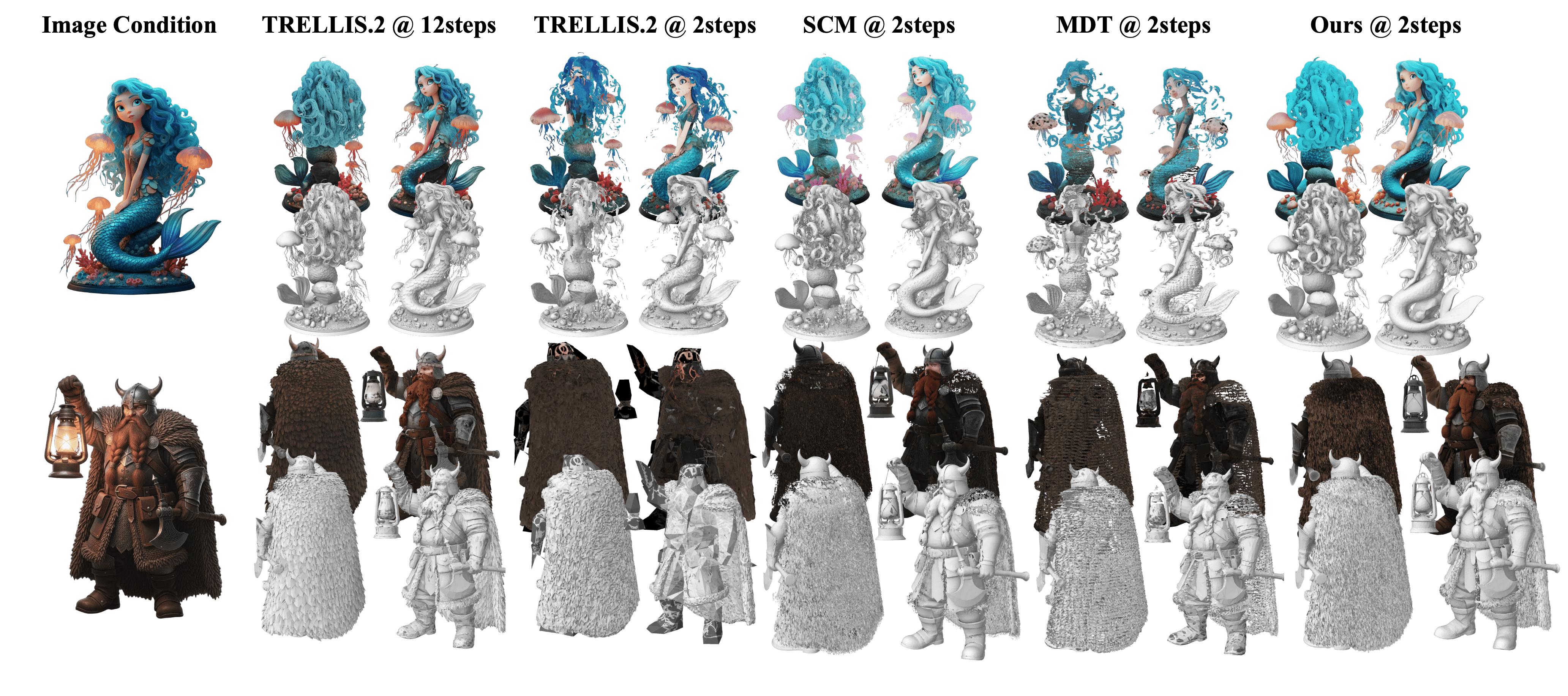}
  \caption{Qualitative comparison on TRELLIS.2 demonstration assets. From left: input image, the 12 and 2-step teacher, SCM @2, MDT @2, and ours @2; each block shows the PBR render and untextured mesh. MDT fractures the geometry, SCM leaves surface artifacts, and ours is closest to the teacher.}
  \label{fig:main}
\end{figure*}

Table~\ref{tab:main} reports the main comparison. Among two-step configurations, our student attains the best Inception and DINOv2 Fréchet distances and the highest ULIP,
improving on the two-step teacher, on the reimplemented marginal-data-transport baseline, and on the self-consistency-only variant, and it matches the twelve-step teacher on ULIP. Figure~\ref{fig:main} shows the same ordering qualitatively, where MDT fractures the geometry and SCM produces less complete geometry than ours. The four-step student narrows most of the remaining Fréchet-distance gap. 
Table~\ref{tab:speed} reports timing: distilling the shape sampler to two steps cuts
shape sampling from $5.29$\,s to $1.18$\,s per asset ($4.5\times$) and end-to-end
image-to-shape generation from $11.51$\,s to $7.55$\,s per asset ($1.5\times$); the
end-to-end factor is bounded by the shared prefix and the unchanged texture stage.

\begin{table*}[t]
  \centering
  \small
  \renewcommand{\arraystretch}{0.95}
  \setlength{\tabcolsep}{4pt}
  \begin{tabular}{lccccc}
    \toprule
    Method & Steps & Time (s) & FD-Incep\,$\downarrow$ & FD-DINOv2\,$\downarrow$ & ULIP-I\,$\uparrow$ \\
    \midrule
    \multicolumn{6}{l}{\textit{Prior methods (cited; Toys4K subset, each method's own protocol)}}\\
    LGM$^{*}$~\citep{tang2024lgm}              & --          & 5  & 26.31 & 322.71 & -- \\
    3DTopia-XL$^{*}$~\citep{chen20243dtopiaxl} & 25          & 5  & 37.68 & 437.37 & -- \\
    Ln3Diff$^{*}$~\citep{lan2024ln3diff}       & 250         & 8  & 26.61 & 357.93 & -- \\
    TRELLIS$^{*}$~\citep{xiang2024trellis}     & 25$\times$2 & -- & 9.35  & 67.21  & -- \\
    \midrule
    \multicolumn{6}{l}{\textit{Our pipeline (TRELLIS.2 backbone; full Toys4K; fixed textured-mesh protocol)}}\\
    Teacher @12 (upper bound)                     & 12 & 11.51 & 12.56 & 106.66 & 38.45 \\
    Teacher @2                                    & 2  & --    & 15.67 & 155.40 & 38.23 \\
    MDT~\citep{zhou2025mdtdist} @2 (reimpl.)      & 2  & --    & 15.98 & 132.80 & 38.04 \\
    SCM @2 (ours, no $\mathcal{L}_{\mathrm{gt}}$) & 2  & --    & 14.76 & 135.02 & 38.43 \\
    \midrule
    Ours @1                                       & 1  & --    & 17.56 & 152.58 & 38.21 \\
    Ours @2                                       & 2  & 7.55  & 14.09 & 127.44 & \textbf{38.48} \\
    Ours @4                                       & 4  & --    & \textbf{13.40} & \textbf{113.56} & 38.49 \\
    \bottomrule
  \end{tabular}
  \caption{Main comparison on Toys4K. ULIP-I is $\times100$; Fréchet distances are set-level. Starred rows are cited from prior work under each method's own protocol and are not directly comparable to the lower block. Times are A800-80GB end-to-end image-to-shape latency where measured.}
  \label{tab:main}
\end{table*}

\begin{table}[ht]
  \centering
  \small
  \renewcommand{\arraystretch}{0.95}
  \setlength{\tabcolsep}{4pt}
  \begin{tabular}{lcc}
    \toprule
    Stage & Teacher & Ours @2 \\
    \midrule
    Shape sampling (s/asset)              & 5.29  & \textbf{1.18} \ ($4.5\times$) \\
    End-to-end image$\to$shape (s/asset)  & 11.51 & \textbf{7.55} \ ($1.5\times$) \\
    \bottomrule
  \end{tabular}
  \caption{Latency on a single A800-80GB GPU.}
  \label{tab:speed}
\end{table}

\subsection{Channel Match Improves the Texture}
\label{sec:cmresults}
Table~\ref{tab:tex} reports the paired render comparison on the demonstration assets.
While the uncalibrated two-step student tracks the teacher, both Channel Match forms improve all metrics. The full match optimizes LPIPS ($0.0802$ to $0.0747$), SSIM ($0.8819$ to $0.8884$), and PSNR ($24.44$ to $24.89$), moving the student closer to the teacher on $73\%$--$88\%$ of assets.

\begin{table}[htbp]
  \centering
  \small
  \renewcommand{\arraystretch}{0.95}
  \setlength{\tabcolsep}{4pt}
  \begin{tabular}{lccc}
    \toprule
    Calibration & LPIPS\,$\downarrow$ & SSIM\,$\uparrow$ & PSNR\,$\uparrow$ \\
    \midrule
    None (two-step student) & 0.0802 & 0.8819 & 24.44 \\
    Std only                & 0.0762 & 0.8860 & 24.72 \\
    Full (mean $+$ std)     & \textbf{0.0747} & \textbf{0.8885} & \textbf{24.85} \\
    \midrule
    \multicolumn{4}{l}{\textit{Fraction of assets moved toward teacher ($\uparrow$)}}\\
    Std only                & 73\% & 75\% & 67\% \\
    Full (mean $+$ std)     & 73\% & 88\% & 73\% \\
    \bottomrule
  \end{tabular}
  \caption{Channel Match on the TRELLIS.2 demonstration assets ($n{=}72$), scored
  against the twelve-step teacher's renders under the four-view protocol. Both forms move the two-step student toward the teacher; the full match gives the larger gain on these artifact-heavy assets.}
  \label{tab:tex}
\end{table}

\subsection{Ablations}
\label{sec:cmablate}

\paragraph{Form of the calibration.}
Table~\ref{tab:cmform} isolates the calibration form on the two-step student over the full Toys4K set. On these geometrically simple assets, where the artifact is rare,
matching only the standard deviation is the safe choice: it slightly improves the Inception Fréchet distance, whereas the full match, which also moves the mean, is a touch worse than doing nothing. This is the opposite of the demonstration assets in
Table~\ref{tab:tex}, where the artifact is common and the full match helps most. The two observations agree: the full match helps most where the collapse is severe (the artifact-heavy demonstration assets), and the std-only match is the safer choice where it is mild (the simple Toys4K assets). The same std match lowers both Fréchet distances for the naively two-step-sampled teacher (Table~\ref{tab:cmform}, lower block), so the shift is distributional rather than specific to the distilled student.

\begin{table}[htbp]
  \centering
  \small
  \renewcommand{\arraystretch}{0.95}
  \setlength{\tabcolsep}{4pt}
  \begin{tabular}{lccc}
    \toprule
    Calibration & ULIP-I\,$\uparrow$ & FD-Incep\,$\downarrow$ & FD-DINOv2\,$\downarrow$ \\
    \midrule
    \multicolumn{4}{l}{\textit{Student @2}}\\
    None                & 38.48 & 14.09 & 127.44 \\
    Std only            & 38.46 & \textbf{13.88} & \textbf{126.88} \\
    Full (mean $+$ std) & 38.38 & 14.16 & 127.88 \\
    \midrule
    \multicolumn{4}{l}{\textit{Teacher @2}}\\
    None                & 38.23 & 15.67 & 155.40 \\
    Std only            & 38.28 & \textbf{15.21} & \textbf{153.12} \\
    Full (mean $+$ std) & 38.35 & 15.66 & 155.03 \\
    \bottomrule
  \end{tabular}
  \caption{Calibration form at two steps on the full Toys4K set, for both the distilled student and the naively teacher.}
  \label{tab:cmform}
\end{table}

\paragraph{Adaptive gate.}
As an upper bound, an oracle that keeps, per asset, whichever of the uncalibrated or calibrated latent is closer to the ground truth lowers the two-step Inception Fréchet distance of Table~\ref{tab:main} from $14.09$ to $13.53$ while calibrating about half of the assets, so per-asset gating has headroom. The deployable gate of Equation~\ref{eq:collapse} approximates this without ground truth, applying the calibration only when the collapse score exceeds a threshold $\tau_S$. Table~\ref{tab:gate} sweeps $\tau_S$ on the demonstration assets: LPIPS, SSIM, and PSNR are essentially flat once the match is applied to at least $75\%$ of assets, so the result is insensitive to $\tau_S$ in this range. The default $\tau_S{=}0.20$ gives the best LPIPS and PSNR, and always-on calibration is a safe fallback. The oracle bound above is on Toys4K FD, so it is not directly comparable to this sweep; closing the gap to it is left open.

\begin{table}[htbp]
  \centering
  \small
  \renewcommand{\arraystretch}{0.95}
  \setlength{\tabcolsep}{4pt}
  \begin{tabular}{lcccc}
    \toprule
    $\tau_S$ & applied & LPIPS\,$\downarrow$ & SSIM\,$\uparrow$ & PSNR\,$\uparrow$ \\
    \midrule
    $-\infty$ (always)   & 100\% & 0.0747 & \textbf{0.8885} & 24.85 \\
    $0.20$ (default)     & 77\%  & \textbf{0.0742} & 0.8884 & \textbf{24.89} \\
    $0.27$               & 50\%  & 0.0761 & 0.8862 & 24.72 \\
    $+\infty$ (never)    & 0\%   & 0.0802 & 0.8819 & 24.44 \\
    \bottomrule
  \end{tabular}
  \caption{Sensitivity to the gating threshold $\tau_S$ on the demonstration assets
  ($n{=}72$, full mean$+$std match), scored by paired LPIPS/SSIM/PSNR against the
  twelve-step teacher under the four-view protocol.}
  \label{tab:gate}
\end{table}
\paragraph{Losses.}
Removing the ground-truth velocity loss and keeping only the self-consistency loss
(the SCM row of Table~\ref{tab:main}) raises the two-step Inception Fréchet distance
from $14.09$ to $14.76$ and the DINOv2 Fréchet distance from $127.44$ to $135.02$, and
lowers ULIP, confirming that the two losses are complementary: self-consistency makes
the field step-agnostic while the velocity loss anchors its values. Removing instead
the self-consistency loss leaves a correct but not step-agnostic field that degrades
sharply at two steps, the standard failure of few-step flow sampling without
distillation.

\subsection{Latent Drift and Sampling Steps}
\label{sec:drift}
Table~\ref{tab:drift} quantifies the shift of Section~\ref{sec:cm} over Toys4K.
The latent is per-channel standardized, so $\hat\sigma_c=1$ and the drift is one-sided: sampled channels fall below the training std. The collapse score $S$ (Eq.~\ref{eq:collapse}) sums exactly this deficit over channels, so it is a direct
scalar measure of the shift. Both $S$ and the fraction of channels with sampled std below $0.8$ grow monotonically as steps shrink, and the four-step student matches the teacher's level, consistent with its being the strongest few-step
configuration in Table~\ref{tab:main}. The intervention (Table~\ref{tab:tex}) shows the shift causes the artifact: matching the statistics removes it, and the monotonic trend here is consistent with the same account.

\begin{table}[htbp]
  \centering
  \small
  \setlength{\tabcolsep}{6pt}
  \begin{tabular}{lccc}
    \toprule
    Config & Steps & Collapse $S$\,$\downarrow$ & Collapsed chans.\,$\downarrow$ \\
    \midrule
    Teacher & 12 & 0.150 & 16.8\% \\
    Student & 4  & 0.146 & 17.5\% \\
    Student & 2  & 0.174 & 22.4\% \\
    Student & 1  & 0.207 & 28.4\% \\
    \bottomrule
  \end{tabular}
  \caption{Latent drift versus sampling steps on Toys4K. Collapse score $S$ (Eq.~\ref{eq:collapse}) is the mean per-channel std deficit; it and the fraction of channels with sampled std below 
$0.8$.}
  \label{tab:drift}
\end{table}

\subsection{JVP Kernel Cost and Stability}
\label{sec:jvpcost}

Table~\ref{tab:jvpnaive} profiles the fused kernel on an A800-80GB GPU. The forward-mode tangent adds training-time cost only: inference remains unaffected. The essential property is memory rather than latency. Against a naive \texttt{torch.func.jvp} baseline, our kernel is $10.9\times$ faster ($41.7$ vs $454.4$,ms) and $O(N)$ rather than $O(N^2)$ in memory, so where the naive path needs $67.6$,GB and runs out of memory at the $N{=}30{,}000$ production length, our kernel still runs. This $O(N)$ memory is what lets the continuous objective fit on the high-resolution sparse latent. The tangent is numerically stable: over $58{,}952$ tokens we saw no NaN or Inf, and the cap of Equation~\ref{eq:tannorm} at $\kappa=4$ roughly halves the maximum tangent norm ($175.0$ to $82.3$) while leaving the median near $21$, so it stabilizes the objective without distorting typical tangents.
\begin{table}[htbp]
  \centering
  \small
  \setlength{\tabcolsep}{4pt}
  \renewcommand{\arraystretch}{1.0}
  \begin{tabular}{@{}lcc@{}}
    \toprule
    Method & Lat.\,(ms)\,$\downarrow$ & VRAM\,(GB)\,$\downarrow$ \\
    \midrule
    FlashAttn foward (no JVP)$^{\dag}$ & 4.8 & --- \\
    \midrule
    Naive foward            & 114.4 & 27.8 \\
    Naive foward\,$+$\,JVP  & 454.4 & 67.6 \\
    \textbf{Triton JVP (ours)} & \textbf{41.7} & \textbf{$\sim$0.2} \\
    \bottomrule
  \end{tabular}
  \caption{Fused JVP attention vs.\ the naive PyTorch baseline (A800-80GB, bf16,
    $N{=}15{,}000$). $^{\dag}$FlashAttention (forward only, no tangent) is a speed \emph{ceiling}, not a JVP baseline.}
  \label{tab:jvpnaive}
\end{table}

\section{Limitations}
We distill only the shape stage, so the end-to-end speedup is bounded by the frozen texture sampler, and distilling it is a direct next step. Channel Match corrects the per-channel distribution shift and removes the resulting texture artifacts, but it does not synthesize new geometric detail, so the four-step student stays strongest on geometry-sensitive metrics; being a per-channel affine map, it also ignores inter-channel and spatial structure, which a full-covariance or grouped variant could capture. The collapse-score gate already recovers most of the available gain, and narrowing the remaining margin to the ground-truth oracle is a promising direction. Extending the same calibration to the texture stage is a natural next question we leave open.

\section{Conclusion}
We distilled the high-resolution shape stage of TRELLIS.2 into a two-step student, using a continuous self-consistency loss, made tractable on the sparse high-resolution latent by a custom JVP FlashAttention kernel, together with an analytic ground-truth velocity loss. At two steps the student reaches geometry comparable to the many-step teacher, matching it on ULIP with a small residual Fréchet-distance gap, at $4.5\times$ faster shape sampling, improving on a marginal-data-transport baseline and a self-consistency-only variant. We further identified a few-step failure mode, an out-of-distribution shape latent that drives the frozen texture model to produce artifacts, and resolved it with Channel Match, a training-free per-channel calibration on the shape latent that removes the texture artifacts and moves the student's appearance toward the teacher, and that reduces the same degradation for a naively few-step-sampled teacher, without retraining either model.

\bibliographystyle{plainnat}
\bibliography{aaai2027}
\clearpage
\appendix
\section{JVP FlashAttention for Sparse Latents}

This appendix gives the full implementation of the fused forward-mode (JVP)
attention kernel used to train the continuous self-consistency objective, and its
integration with PyTorch forward-mode autodiff. We first restate why a custom kernel
is needed, derive the forward-mode tangent of softmax attention, describe the fused
online-softmax algorithm that evaluates it in $O(N)$ memory, and finally give the
autograd bridge that exposes the kernel to \texttt{torch.autograd.forward\_ad}.

\subsection{Why a Custom Kernel}
The continuous self-consistency loss requires the total time derivative of the
student velocity along a teacher trajectory,
\begin{equation}
  g(x_t,t) = \partial_t v_\theta + (\nabla_x v_\theta)\,\bar v_\phi ,
  \label{eq:tangent-supp}
\end{equation}
which is a Jacobian-vector product (JVP): a single forward-mode pass through the
backbone with the latent tangent seeded by the teacher velocity $\bar v_\phi$ and the
time tangent seeded by $\mathrm{d}(1000\,t)/\mathrm{d}t=1000$. Forward-mode
differentiation must propagate a tangent through \emph{every} layer, including
attention. Stock FlashAttention~\citep{dao2022flashattention} returns only the primal
output and exposes no tangent, and a naive PyTorch attention that supports forward-mode
AD (\texttt{torch.func.jvp}) materializes the $N\times N$ score matrix, whose $O(N^2)$
memory is infeasible at the $1024^3$ sparse-latent token counts. We therefore implement
a fused kernel that returns the primal output and its tangent in one online-softmax pass
at the memory cost of standard attention.

\subsection{Forward-Mode Tangent of Softmax Attention}
Let $S=QK^\top/\sqrt{d}$ be the scores, $P=\mathrm{softmax}(S)$ the row-normalized
weights, and $O=PV$ the output. Given input tangents $(\dot Q,\dot K,\dot V)$, the score
tangent is
\begin{equation}
  \dot S = \big(\dot Q\,K^\top + Q\,\dot K^\top\big)\big/\sqrt{d}.
\end{equation}
The softmax Jacobian gives, per query row $i$,
\begin{equation}
  \dot P_{ij} = P_{ij}\Big(\dot S_{ij} - \textstyle\sum_{j'} P_{ij'}\dot S_{ij'}\Big),
\end{equation}
and since $O=PV$,
\begin{equation}
  \dot O = \dot P\,V + P\,\dot V .
\end{equation}
Collecting terms per row and defining the four row-wise quantities
\begin{align}
  O_i &= \textstyle\sum_j P_{ij} V_j, &
  A_i &= \textstyle\sum_j P_{ij} \dot V_j,\\
  B_i &= \textstyle\sum_j P_{ij} \dot S_{ij} V_j, &
  D_i &= \textstyle\sum_j P_{ij} \dot S_{ij},
\end{align}
the output tangent reduces to the compact form
\begin{equation}
  \dot O_i \;=\; A_i + B_i - D_i\,O_i .
  \label{eq:jvp-compact}
\end{equation}
All four quantities are row sums over keys, so they admit the same streaming,
$O(N)$-memory accumulation as the FlashAttention denominator.

\subsection{Fused Online-Softmax Algorithm}
The kernel keeps FlashAttention's numerically stable online softmax (running row max
$m_i$ and denominator $l_i$) and accumulates the four unnormalized quantities
$(\tilde O_i,\tilde A_i,\tilde B_i,\tilde D_i)$ alongside it, rescaling all of them by
the same $\exp(m_i^{\text{old}}-m_i^{\text{new}})$ factor whenever the running max
updates. After the key loop it normalizes once,
\begin{equation}
  O = \tilde O/l,\qquad
  \dot O = (\tilde A+\tilde B)/l - (\tilde D/l)\,O,
\end{equation}
which is exactly Equation~\ref{eq:jvp-compact}. The $N\times N$ scores and their
tangent are never materialized: each key block contributes to the accumulators and is
discarded, so peak memory matches standard attention plus the tangent input/output
tensors. Listing~\ref{lst:kernel} shows the inner loop; the kernel is autotuned over
block sizes keyed on the head dimension, runs on variable-length packed sequences via
\texttt{cu\_seqlens} (matching the sparse latent's per-sample token counts), loads in
bfloat16, and accumulates in fp32.

\begin{listing}[t]
\caption{Fused forward\,+\,JVP attention (Triton, inner loop).}
\label{lst:kernel}
\begin{lstlisting}[language=Python]
# one program := (query block pid_m, batch*head)
m_i, l_i = -inf, 0.0            # online-softmax max, denom
O_i = A_i = B_i = 0.0          # [BLOCK_M, d]  primal + tangents
D_i = 0.0                      # [BLOCK_M]
for start_n in range(0, seqlen_k, BLOCK_N):
  # load q,k,v and tangents tq,tk,tv (fp32)
  S  = dot(q, k) * sm_scale               # scores
  tS = (dot(tq, k) + dot(q, tk)) * sm_scale  # score tangent
  S  = mask(S, -inf); tS = mask(tS, 0.0)  # padding/causal
  m_new = maximum(m_i, max(S, axis=1))
  P  = exp(S - m_new)                      # unnormalized
  a  = exp(m_i - m_new)                    # rescale old state
  l_i = a*l_i + sum(P, axis=1)
  O_i = a*O_i + dot(P,  v)                 # Sum P v
  A_i = a*A_i + dot(P,  tv)                # Sum P tv
  PdS = P * tS
  B_i = a*B_i + dot(PdS, v)                # Sum P tS v
  D_i = a*D_i + sum(PdS, axis=1)           # Sum P tS
  m_i = m_new
O  = O_i / l_i
tO = (A_i + B_i) / l_i - (D_i / l_i) * O   # Eq. (jvp-compact)
store(O); store(tO)
\end{lstlisting}
\end{listing}

\subsection{Integration with Forward-Mode Autodiff}
The kernel is exposed to PyTorch through a custom \texttt{torch.autograd.Function} whose
\texttt{forward} calls the original FlashAttention (primal only), whose \texttt{backward}
keeps standard reverse-mode available (re-running FlashAttention under
\texttt{enable\_grad}), and whose \texttt{jvp} static method invokes the fused Triton
kernel to return the output tangent. During the tangent pass the backbone's
variable-length attention entry points are monkey-patched to this Function by a context
manager and restored on exit, so only the forward-mode pass routes through the kernel and
the rest of the pipeline is untouched (Listing~\ref{lst:bridge}). At the call site the
training step seeds the duals---the latent tangent with $\bar v_\phi$ and the scaled-time
tangent with the constant $1000$---inside a \texttt{fwAD.dual\_level()} under the patch,
then unpacks the dual of the output to obtain $g$ of Equation~\ref{eq:tangent-supp}.

\begin{listing}[t]
\caption{Autograd bridge and attention patch.}
\label{lst:bridge}
\begin{lstlisting}[language=Python]
class FlashAttnVarlenFuncJVP(torch.autograd.Function):
  @staticmethod
  def forward(ctx, q,k,v, cu_q,cu_k, mq,mk):
    ctx.save_for_backward(q,k,v); ...
    return flash_attn_varlen_func(q,k,v, ...)  # primal
  @staticmethod
  def backward(ctx, g):                        # reverse-mode kept
    with torch.enable_grad():
      out = flash_attn_varlen_func(q_d,k_d,v_d, ...)
      return torch.autograd.grad(out, [q_d,k_d,v_d], g)
  @staticmethod
  def jvp(ctx, dq,dk,dv, *a):                   # <- forward-mode
    out, tout = empty_like(q), empty_like(q)
    _attn_fwd_jvp_kernel[grid](q,k,v, dq,dk,dv, out,tout, ...)
    return tout

@contextlib.contextmanager
def patch_trellis_attention_for_jvp():
  orig = flash_attn.flash_attn_varlen_func
  flash_attn.flash_attn_varlen_func = \
     lambda q,k,v,*a,**kw: FlashAttnVarlenFuncJVP.apply(q,k,v,*a)
  try:     yield
  finally: flash_attn.flash_attn_varlen_func = orig
\end{lstlisting}
\end{listing}

\subsection{Cost and Numerical Stability}
The fused kernel keeps forward-mode differentiation at the memory cost of standard
attention: it is $O(N)$ in the token count rather than $O(N^2)$, so it runs at the
$N{=}30{,}000$ production length where the naive \texttt{torch.func.jvp} baseline needs
tens of gigabytes and runs out of memory (main paper, JVP-cost table). The overhead is
training-time only---inference uses the plain student and never evaluates the tangent.
The per-token tangent $g$ is stabilized by the smooth magnitude-preserving cap of the
main paper before it enters the self-consistency target; over the training tokens we
profiled we observed no NaN or Inf, and the cap bounds the maximum tangent norm while
leaving the median essentially unchanged.

\section{Additional Qualitative Results}
\label{sec:qualitative}
\begin{figure*}[t]
  \centering
  \includegraphics[width=1\textwidth]{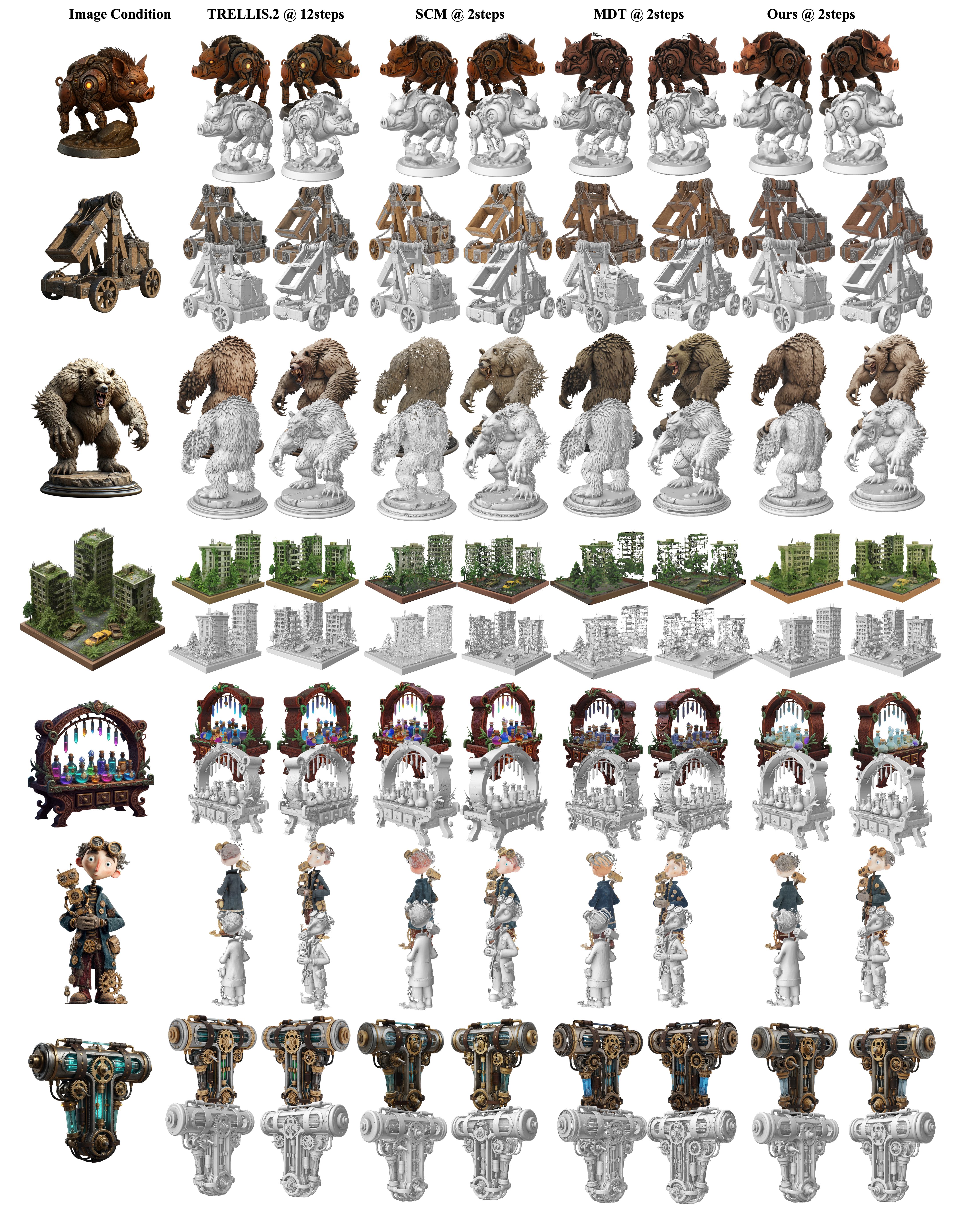}
  \caption{Additional qualitative comparison on TRELLIS.2 demonstration assets. From left to right: input image, the 12-step teacher, SCM @2, MDT @2, and ours @2. Each block displays the PBR render and the untextured mesh. MDT fractures the geometry, SCM leaves surface artifacts, and our method produces results closest to the teacher.}
  \label{fig:sp1}
\end{figure*}

In this section, we present additional qualitative comparisons to further demonstrate the effectiveness of our proposed distillation framework and Channel Match. Figure~\ref{fig:sp1} provides extended visual comparisons on the TRELLIS.2 demonstration assets. We compare our two-step student model against the twelve-step teacher model, as well as the Self-Consistency Match (SCM) and Marginal-Data-Transport (MDT) baselines. Consistent with the main text, our approach recovers geometric details and texture quality that closely approximate the many-step teacher. In contrast, the MDT baseline frequently fractures the underlying geometry, while the SCM baseline preserves geometric completeness but introduces noticeable surface artifacts due to the uncalibrated shape latents.
\section{Channel Match on TRELLIS.1}
\label{sec:trellis1}
\begin{figure*}[t]
  \centering
  \includegraphics[width=0.9\textwidth]{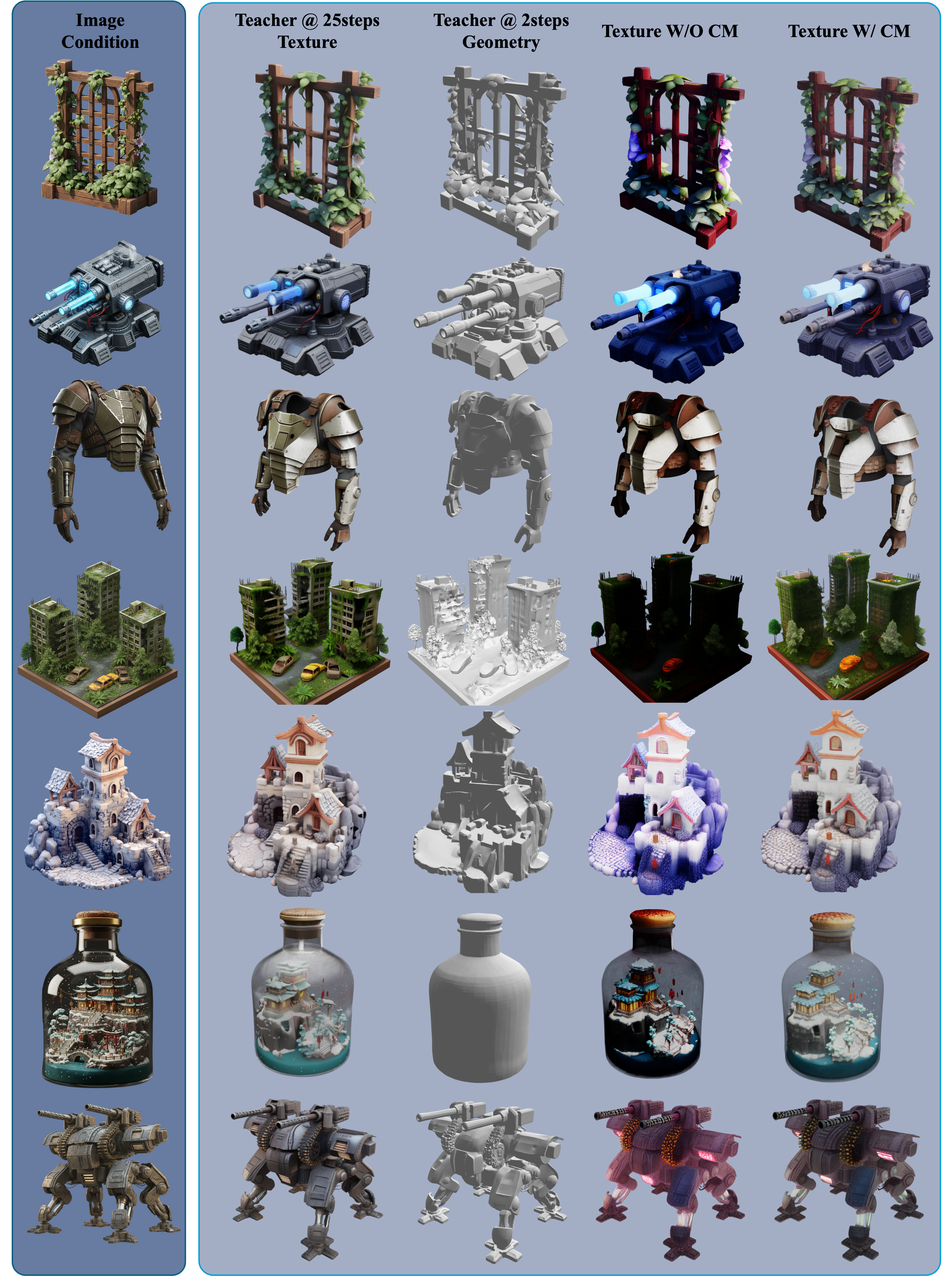}
  \caption{The effect of Channel Match on the TRELLIS.1 architecture (variable voxel resolution). From left to right: input image; the untextured mesh and render of the two-step student without CM; and the mesh and render with CM. Although TRELLIS.1 does not exhibit the severe dark artifacts seen in TRELLIS.2, the few-step sampling still degrades the overall texture quality. Applying CM effectively restores the texture fidelity by calibrating the shifted shape latent.}
  \label{fig:trellis1_sp3}
\end{figure*}

To evaluate the generalization of our latent calibration method, we apply Channel Match to the earlier TRELLIS.1 architecture. Unlike TRELLIS.2, TRELLIS.1 operates without a fixed voxel resolution. As illustrated in Figure~\ref{fig:trellis1_sp3}, while the distribution shift induced by few-step sampling in TRELLIS.1 does not manifest as the severe dark or blotchy surface artifacts observed in TRELLIS.2, it nonetheless causes a significant overall degradation in texture quality. Applying our training-free Channel Match accurately restores the per-channel statistics of the few-step shape latents. This correction successfully recovers the texture fidelity produced by the downstream appearance stage, confirming that sampling-induced latent shift is a general phenomenon across different structured-latent pipelines and can be mitigated by our calibration method.
\section{High-Resolution Generation on TRELLIS.2}
\label{sec:highres}
\begin{figure*}[t]
  \centering
  \includegraphics[width=0.6\textwidth]{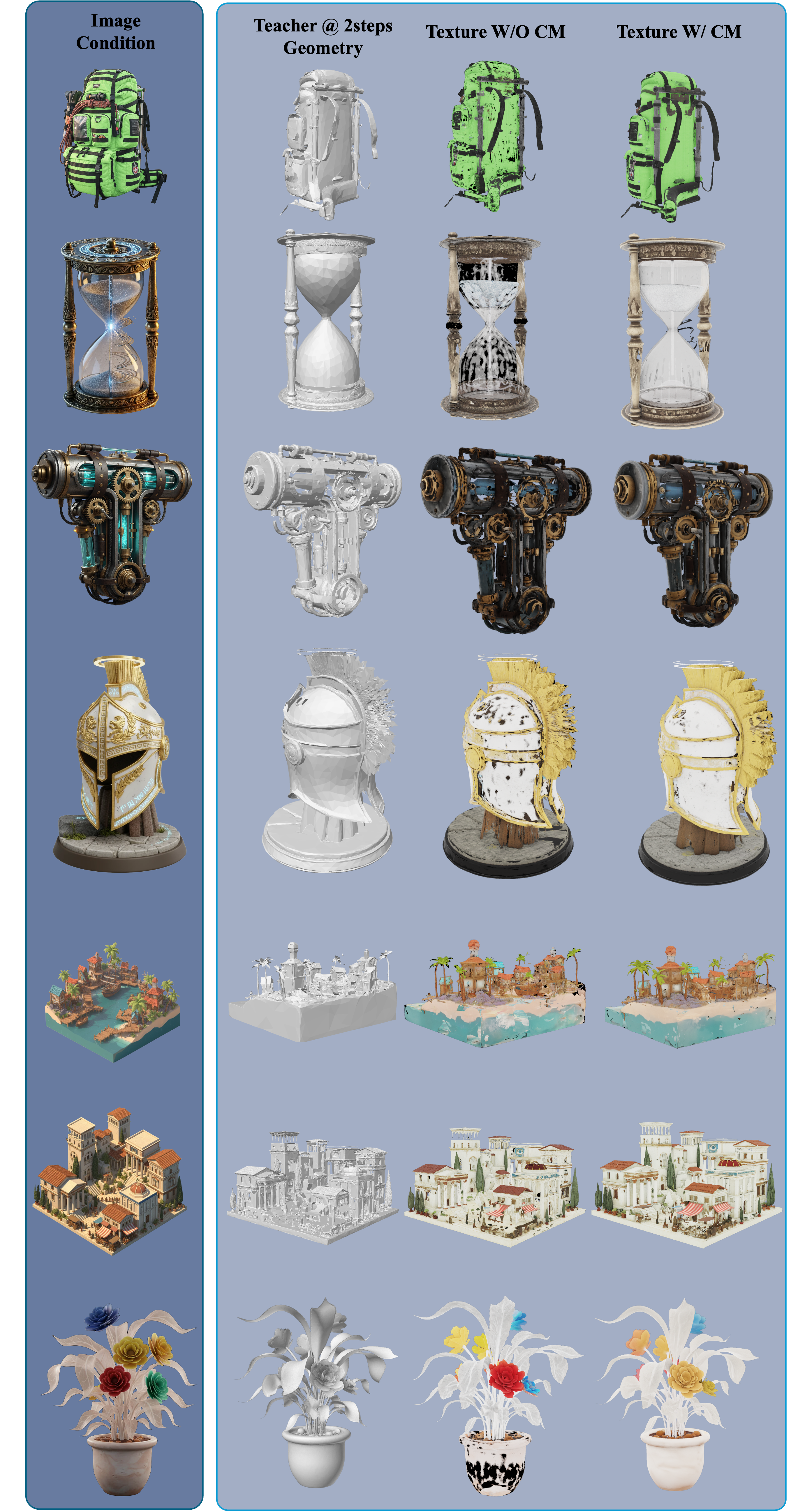}
  \caption{High-resolution ($1536^3$) shape generation on TRELLIS.2. The figure demonstrates the few-step distribution shift and its correction by Channel Match (CM). From left to right: input image; the untextured mesh and PBR render of the two-step student without CM, where the shifted shape latent causes the frozen texture model to produce faulty surface materials (e.g., dark patches); and the corresponding mesh and render with CM applied. Calibrating the latent removes the texture artifacts while maintaining the high-resolution geometry. We consider samples with relatively simple geometry to isolate the effect of latent distribution shift from geometry reconstruction errors, demonstrating that CM effectively corrects the shift at high resolutions.}
  \label{fig:sp2_1536}
\end{figure*}

We further validate the robustness of Channel Match at a higher generation resolution. Specifically, we evaluate the few-step shape generation on TRELLIS.2 scaled to a $1536^3$ grid resolution. As shown in Figure~\ref{fig:sp2_1536}, the few-step distribution shift persists at this higher resolution, causing the geometry latent to drift away from the training distribution expected by the frozen PBR texture model and resulting in incorrect material synthesis. The application of Channel Match effectively corrects this shift, eliminating the texture artifacts and restoring the rendering quality without altering the underlying high-resolution geometry. This confirms that our calibration strategy scales reliably to higher-dimensional sparse latents.

\end{document}